\documentclass[default,iicol]{sn-jnl}

\jyear{2021}%

\theoremstyle{thmstyleone}%
\theoremstyle{thmstyletwo}%

\theoremstyle{thmstylethree}%
\usepackage{pifont}
\usepackage{color}

\usepackage{makecell}
\usepackage{multirow}
\usepackage{booktabs}
\usepackage{enumerate}

\newcommand{\PreserveBackslash}[1]{\let\temp=\\#1\let\\=\temp}
\newcolumntype{C}[1]{>{\PreserveBackslash\centering}p{#1}}
\newcolumntype{R}[1]{>{\PreserveBackslash\raggedleft}p{#1}}
\newcolumntype{L}[1]{>{\PreserveBackslash\raggedright}p{#1}}

\newcommand{\tabincell}[2]{\begin{tabular}{@{}#1@{}}#2\end{tabular}}

\def\red#1{\textcolor{red}{#1}}
\def\blue#1{\textcolor{blue}{#1}}

\def\comment#1{{}}
\def\eg{{\em e.g.}}
\def\ie{{\em i.e.}}
\def\etc{{\em etc}}
\def\etal{{\em et al.}}

\begin{document}

\title[Article Title]{PIDray: A Large-scale X-ray Benchmark for Real-World Prohibited Item Detection}


\author[1]{Libo Zhang}\email{libo@iscas.ac.cn}\equalcont

\author[1,2]{Lutao Jiang}\email{lutao2021@iscas.ac.cn}\equalcont

\author*[1,2]{Ruyi Ji}\email{ruyi2017@iscas.ac.cn}

\author[3]{Heng Fan}\email{heng.fan@unt.edu}

\affil[1]{\orgname{Institute of Software Chinese Academy of Sciences}, \orgaddress{\country{China}}}

\affil[2]{\orgname{University of Chinese Academy of Sciences}, \orgaddress{\country{China}}}

\affil[3]{\orgname{Department of Computer Science and Engineering, University of North Texas}, \orgaddress{\country{USA}}}


\abstract{
	Automatic security inspection relying on computer vision technology is a challenging task in real-world scenarios due to many factors, such as intra-class variance, class imbalance, and occlusion. Most previous methods rarely touch the cases where the prohibited items are deliberately hidden in messy objects because of the scarcity of large-scale datasets, hindering their applications. To address this issue and facilitate related research, we present a large-scale dataset, named \textbf{PIDray}, which covers various cases in real-world scenarios for prohibited item detection, especially for deliberately hidden items. In specific, PIDray collects $124,486$ X-ray images for $12$ categories of prohibited items, and each image is manually annotated with careful inspection, which makes it, to our best knowledge, to largest prohibited items detection dataset to date. Meanwhile, we propose a general divide-and-conquer pipeline to develop baseline algorithms on PIDray. Specifically, we adopt the tree-like structure to suppress the influence of the long-tailed issue in the PIDray dataset, where the first course-grained node is tasked with the binary classification to alleviate the influence of head category, while the subsequent fine-grained node is dedicated to the specific tasks of the tail categories. Based on this simple yet effective scheme, we offer strong task-specific baselines across object detection, instance segmentation, and multi-label classification tasks and verify the generalization ability on common datasets (\eg, COCO and PASCAL VOC). Extensive experiments on PIDray demonstrate that the proposed method performs favorably against current state-of-the-art methods, especially for deliberately hidden items. Our benchmark and codes will be released at \url{https://github.com/lutao2021/PIDray}.
}




\maketitle

\section{Introduction}\label{sec1}

Security inspection is tasked with checking packages against specific criteria and reveals any potential risks to ensure public safety, which is widely applied in real-world scenarios, such as public transportation and sensitive departments. In practice, there is an ever-increasing demand for inspectors to monitor the scanned X-ray images generated by the security inspection machine to specify potentially prohibited items, such as guns, ammunition, explosives, corrosive substances, and toxic and radioactive substances. But unfortunately, it is highly challenging for inspectors to localize prohibited items hidden in messy objects accurately and efficiently, which poses a great threat to safety.

\begin{figure}[!t]
	\centering
	\includegraphics[width=\linewidth]{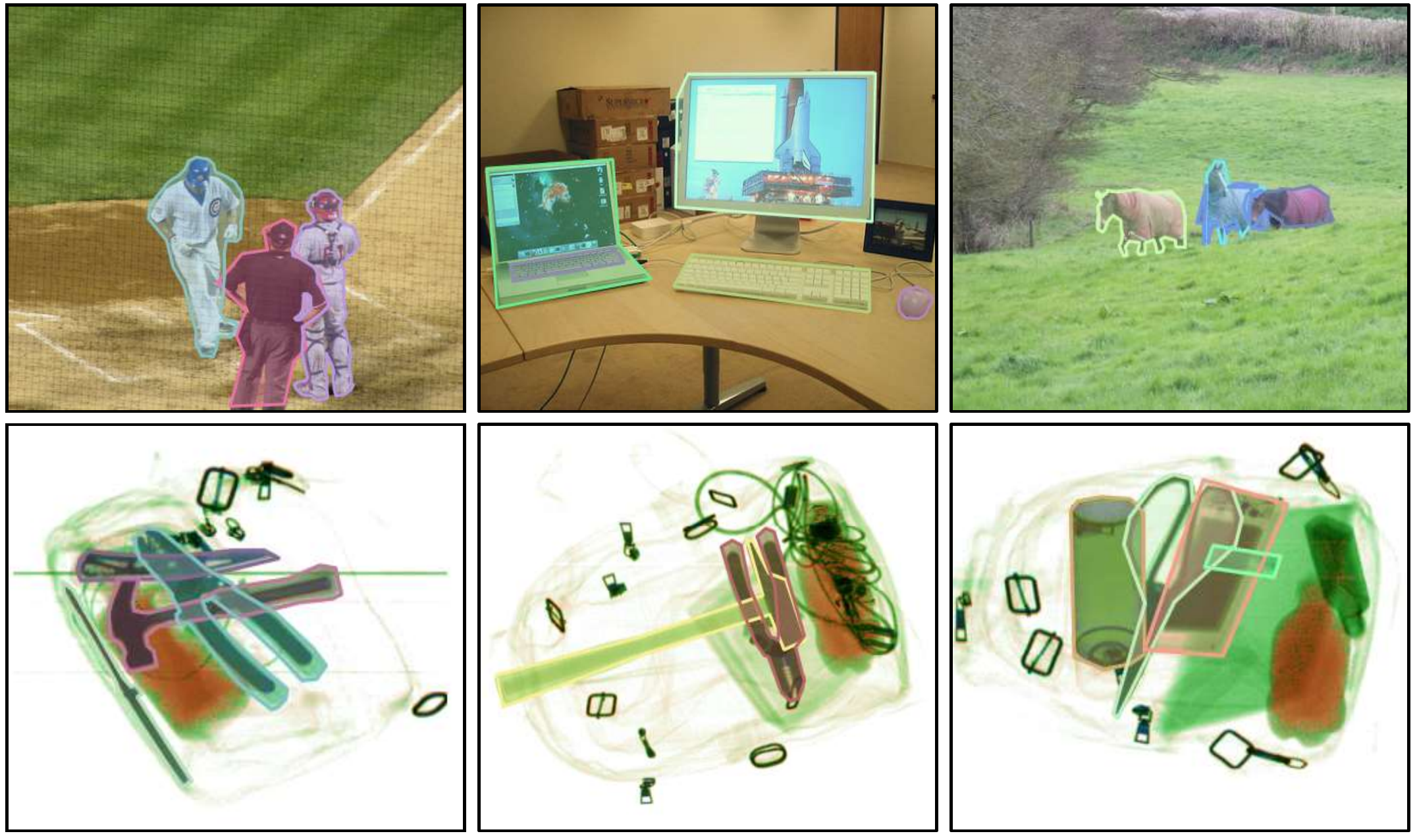}
	\caption{Comparisons between the natural image (the 1-st row) and X-ray image (the 2-nd row).}
	\label{fig:overlapping}
\end{figure}

Deep learning technologies have sparked tremendous progress in computer vision community \cite{ren2015faster,liu2016ssd, tian2019fcos, DBLP:conf/cvpr/JiWZDWZLH20, DBLP:conf/eccv/JiDZWWZHL20, DBLP:conf/eccv/LiDZWLWZ20, DBLP:conf/mm/CaiDZWWWL20}, which makes it possible to inspect prohibited items automatically. The security inspectors demand to quickly identify the locations and categories of prohibited items relying on computer vision technology. Most of the previous object detection algorithms are well-designed to detect objects in natural images, which are not optimal for detection in X-ray images due to the following factors. Firstly, X-rays have strong penetrating power, and different materials in the object absorb X-rays to different degrees, resulting in different colors. Secondly, the contours of the occluder and the occluded objects in the X-ray are mixed together. As shown in Fig.~\ref{fig:overlapping}, compared with natural images, X-ray images present a quite different appearance and edges of objects and background, which brings new challenges in appearance modeling for X-ray detection. To advance the developments of prohibited item detection in X-ray images, some recent attempts devote to construct security inspection benchmarks \cite{mery2015gdxray,akcay2017evaluation,akcay2018using,miao2019sixray,wei2020occluded}. However, most of them fail to meet the requirements in real-world applications for three reasons. (1) Existing datasets are characterized by small volumes and very few categories of prohibited items (\eg, \textit{knife}, \textit{gun} and \textit{scissors}). For example, some common prohibited items such as \textit{powerbank}, \textit{lighter} and \textit{sprayer} are not involved. (2) Some real-world scenarios require a high security level based on accurate predictions of masks and categories of prohibited items. The image-level or bounding-box-level annotations in previous datasets are not sufficient to train algorithms in such scenarios. (3) Detecting prohibited items hidden in messy objects is one of the most significant challenges in security inspection. Unfortunately, few studies are developed towards this goal due to the lack of comprehensive datasets covering such cases.
These challenges urgently require a large-scale prohibited item benchmark as well as an efficient and effective method.

In another hand, we observe that the models trained on the dataset, majority of which are with prohibited items, are error-prone when processing the samples without any prohibited items. We argue that this issue arises from the fact that mainstream training schemes exclude all images without any bounding box by default during the pre-processing stage. Even though no significant influence on general datasets (\eg, COCO \cite{lin2014microsoft} and PASCAL VOC \cite{10.1007/s11263-009-0275-4}) where annotated samples account for the majority, this arrangement of dataset incurs a dilemma in security inspection, as the images with prohibited items are just special cases in the real scenario. Such cases reflect the large gap between artificial datasets and real scenarios. Therefore, for a more stable and robust model in security inspection, the construction of dataset should be in line with the real scenarios as far as possible.

\begin{figure*}[t]
	\centering
	\includegraphics[width=1.0\linewidth]{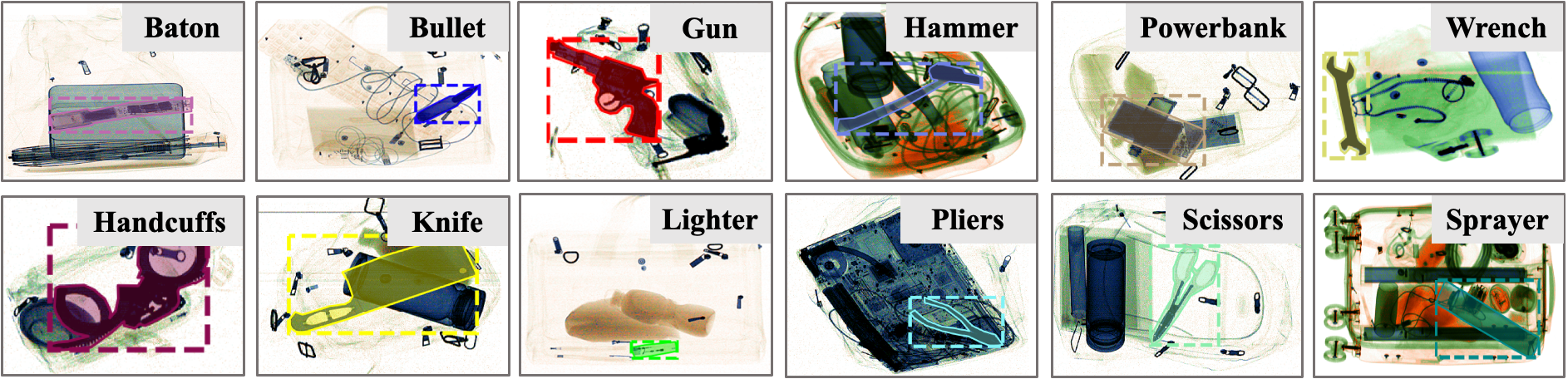}
	\caption{Example images in the PIDray dataset with $12$ categories of prohibited items. Each image is provided with image-level and instance-level annotation. For clarity, we show one category per image.}
	\label{fig:samples}
\end{figure*}
To this end, we present a large-scale prohibited item detection dataset (PIDray) for real-world applications. The PIDray dataset covers $12$ categories of prohibited items in X-ray images. From the exemplars with annotations in Fig.~\ref{fig:samples}, we can observe that each image contains at least one prohibited item with both the bounding box and mask annotations. Notably, for the fine-grained investigation, the test set is well-divided into three parts, \ie, \textit{easy}, \textit{hard} and \textit{hidden}. Particularly, the \textit{hidden} subset focuses on the prohibited items deliberately hidden in messy objects (\eg, change the item shape by wrapping wires). To the best of our knowledge, this is the largest dataset to date for the detection of prohibited items.

Based on the observation that images without prohibited items account for the majority of the proposed dataset, which characterizes the dataset with long-tailed distribution, we propose a divide-and-conquer pipeline, which adopts the tree-like structure to suppress the influence of samples without prohibited items in a course-to-fine manner. Specifically, the sample first passes through the first coarse-grained node to determine whether it contains the prohibited item or not. If true, the sample is fed to the subsequent fine-grained node for the task-specific operations (\eg, detection or segmentation). If not, that means this is a  sample without any prohibited items. The key insight of our method is that the distribution of the proposed dataset enables us to cast such a task as a multi-task learning problem. We perform the binary classification in the first node to balance the head and tail categories in a course-grained manner, and then perform  task-specific operations in the later node in a fine-grained manner.

For object detection and instance segmentation tasks, we utilize such a pipeline to construct a strong baseline on the top of two- or one-stage detectors like Cascade Mask R-CNN \cite{cai2019cascade}, where we also contribute to the FPN structure with the dense attention module. Concretely, we first use both the spatial- and channel-wise attention mechanisms to exploit discriminative features, which is effective to locate the deliberately prohibited items hidden in messy objects. Then we design the dependency refinement module to explore the long-range dependencies within feature map. Extensive experiments on the proposed dataset show that our method performs favorably against the state-of-the-art methods. 

Meanwhile, to fully unleash the potential of the proposed dataset, we establish a multi-label classification task for this benchmark. And we extend the divide-and-conquer pipeline to this domain to alleviate the issue of long-tailed distribution. Specifically, the first coarse-grained node is tasked with the binary classification, filtering out head category without prohibited items. After that, the fine-grained node is dedicated to the multi-label classification of the tail categories with prohibited items. The experiment performance demonstrates that our design is a simple yet effective scheme for the multi-label classification task on the PIDray dataset.

To sum up, the main contributions of this work can be summarized into the following four folds. 
\begin{itemize}
	\item Towards the prohibited item detection in real-world scenarios, we present a large-scale benchmark, \ie, PIDray, formed by $124,486$ images in total. To the best of our knowledge, it is the largest X-ray prohibited item detection dataset to date. Meanwhile, it is the first benchmark dedicated to cases where prohibited items are deliberately hidden in messy objects. 

        \item We provide various tasks besides object detection on the proposed PIDray to fully unleash its the potential in real-world application including segmentation and multi-label classification.
        
	\item We propose the divide-and-conquer pipeline to address the issue of long-tailed distribution in the PIDray dataset,  which adopts the tree-like structure to suppress the influence of samples without prohibited items in a course-to-fine manner.
	\item With our novel divide-and-conquer pipeline, we deliver strong task-specific baselines across object detection, instance segmentation, and multi-label classification tasks on the PIDray dataset and verify its generalization ability on common datasets (\eg, COCO and PASCAL VOC). 
	\item Extensive experiments carried  out on the PIDray dataset and general dataset verify the effectiveness of the proposed methods compared to the state-of-the-art methods.
\end{itemize}

This paper extends an early conference version in~\cite{wang2021towards}. The main new contributions are as follows. \textbf{(1)} We enlarge PIDray by introducing 76,809 new X-ray images without prohibited items to bridge the gap between artifact dataset and natural scenarios. \textbf{(2)} We enrich the applications of PIDray by introducing a new task of multi-label classification, which further unleashes the potential of PIDray. \textbf{(3)} We propose a novel divide-and-conquer pipeline for developing strong task-specific baselines on PIDray to facilitate future research. \textbf{(4)} More thorough experiments are conducted on PIDray with in-depth analysis to show the effectiveness of our approach. \textbf{(5)} Besides the experiments on PIDray, we further verify the generalization ability of our pipeline on common benchmarks (\eg, COCO, PASCAL VOC).


The remainder of this paper is organized as follows. Section \ref{related_work} briefly reviews research directions relevant to our method.  In Section \ref{dataset}, we describe the construction of the PIDray dataset in detail. In Sections \ref{object detection} and \ref{multi-label classification}, the task-specific strong baselines are realized under the guidance of divide-and-conquer pipeline. In Section \ref{experiments}, the extensive experimental results are reported and analyzed, including the comparison between the proposed method and the state-of-the-art approaches, validation of generalization ability on the general datasets, and the comprehensive analysis of the ablation studies. Finally, Section \ref{conclution} draws the conclusions of the proposed method.

\section{Related Work}
\label{related_work}

This section reviews six major research directions closely related to our work, \ie, prohibited item benchmarks, object detection, the attention mechanism, multi-label classification, multi-scale feature fusion and long-tailed distribution issue.

\subsection{Prohibited Items Benchmarks}
Due to discrepancies of penetrating capability, different materials tend to present various colors under X-ray. Such property incurs more challenges in cases where objects are overlapped. Moreover, like natural images, X-ray images are featured of notorious characteristics as well, \eg, intra-class variances, distribution imbalance. Recently, there has been a few datasets collected to advance prohibited item detection investigation. To be concrete, \cite{mery2015gdxray} collects GDXray dataset for nondestructive testing. GDXray is formed by three types of prohibited items: gun, shuriken and razor blade. Without complex background and overlap, it is easy to recognize or detect objects in this dataset. Differing from GDXray, Dbf6  \cite{akcay2017evaluation}, Dbf3 \cite{akcay2018using} and OPIXray \cite{wei2020occluded} cover complicated background and overlapping-data, but unfortunately, the volumes of images and prohibited items are still insufficient. Recently, \cite{Jinyi2019xray} releases a dataset containing $ 32,253 $ X-ray images, of which $ 12,683 $ images include prohibited items. This dataset has $ 6 $ types of items, but none of them are strictly prohibited, such as mobile phones, umbrellas, computers, and keys. \cite{miao2019sixray} provides a large-scale security inspection dataset called SIXray, which covers $ 1,059,231 $ X-ray images with image-level annotation. However, the proportion of images containing prohibited items is very small in the dataset (\ie, only $ 0.84\% $). In addition, there are $ 6 $ categories of prohibited items, but only $ 5 $ categories are annotated. Unlike the aforementioned datasets, we propose a new large-scale security inspection benchmark that contains over $ 47 $K images with prohibited items and $ 12 $ categories of prohibited items with pixel-level annotation. Towards real-world application, we focus on detecting deliberately hidden prohibited items.

\begin{table*}[t]
	\centering
	\caption{Comparison of the dataset statistics with existing X-ray benchmarks. ``Total'' and ``Prohibited'' indicate the number of total images and the images containing prohibited items in the dataset, respectively. \textbf{C}, \textbf{O}, \textbf{I}, and \textbf{M} represent Classification, Object Detection, Instance Segmentation and Multi-label classification respectively. \textbf{S}, \textbf{A}, and \textbf{R} represent Subway, Airport, and Railway Station respectively.}
	\label{table:comparison}
	\resizebox{0.99\textwidth}{!}{
		\begin{tabular}{@{}rccccccccccc@{}}
			\toprule
			      \multirow{2}{*}{Dataset}      & \multirow{2}{*}{Year} & \multirow{2}{*}{Classes} & \multicolumn{2}{@{}c@{}}{Images}  &   \multicolumn{3}{@{}c@{}}{Annotations}   & \multirow{2}{*}{Type} & \multirow{2}{*}{Scene} & \multirow{2}{*}{Task} & \multirow{2}{*}{Avail} \\ 
         \cmidrule(l){4-5} \cmidrule(l){6-8} 
			                                     &                       &                          &     Total      & Prohibited  &   Image    &    Bbox    &    Mask    &                       &                        &                              &                               \\ \midrule
			    GDXray \cite{mery2015gdxray}     &        $2015$         &           $3$            &    $8,150$     &   $8,150$   & \checkmark & \checkmark &     -      &         Real          &           -            &             C/O              &          \checkmark           \\
			Dbf$_{6}$ \cite{akcay2017evaluation} &        $2017$         &           $6$            &    $11,627$    &  $11,627$   & \checkmark & \checkmark &     -      &         Real          &           -            &             C/O              &           \ding{55}           \\
			  Dbf$_{3}$ \cite{akcay2018using}    &        $2018$         &           $3$            &    $7,603$     &   $7,603$   & \checkmark & \checkmark &     -      &         Real          &           -            &             C/O              &           \ding{55}           \\
			   Liu~\etal~\cite{Jinyi2019xray}    &        $2019$         &           $6$            &    $32,253$    &  $12,683$   & \checkmark & \checkmark &     -      &         Real          &           S            &             C/O              &           \ding{55}           \\
			    SIXray \cite{miao2019sixray}     &        $2019$         &           $6$            & $\bf1,059,231$ &   $8,929$   & \checkmark & \checkmark &     -      &         Real          &           S            &             C/O              &          \checkmark           \\
			   OPIXray \cite{wei2020occluded}    &        $2020$         &           $5$            &    $8,885$     &   $8,885$   & \checkmark & \checkmark &     -      &       Synthetic       &           A            &             C/O              &          \checkmark           \\
			                Our PIDray                 &        $2022$         &         $\bf12$          &   $124,486$    & $\bf47,677$ & \checkmark & \checkmark & \checkmark &         Real          &         S/A/R          &           C/O/I/M            &          \checkmark           \\ \bottomrule
		\end{tabular}}
\end{table*}

\subsection{Object Detection}
Object detection is a long-standing problem in the computer vision community. Generally speaking, modern object detectors fall into two groups: two-stage and one-stage detectors.

\textbf{Two-stage Detectors.} R-CNN \cite{girshick2014rich} exemplifies the first research line and proves that CNN can dramatically improve detection performance. However, it is time-consuming that each regional proposal is computed individually in this pipeline. To close this gap, Fast-RCNN \cite{girshick2015fast} utilizes the ROI pooling layer to extract fixed-size features for each proposal from the feature map of the full image. The follower \cite{ren2015faster} introduces an alternative way to replace selective search, which designs the RPN network and derives numerous variants. For example, FPN \cite{lin2017feature} assembles low-resolution features with high-resolution features through a top-down pathway and lateral connections. Mask R-CNN \cite{he2017mask} attaches a mask branch to the Faster-RCNN \cite{ren2015faster} structure to improve detection performance via a multi-task learning scheme. Cascade R-CNN \cite{cai2018cascade} introduces the classic cascade structure into Faster R-CNN \cite{ren2015faster} framework. Libra R-CNN \cite{pang2019libra} develops a simple yet effective strategy to alleviate the issue of data imbalance in the training process.

\textbf{One-stage Detectors.} OverFeat \cite{sermanet2013overfeat} is one of the seminal methods relying on deep learning among one-stage detectors. Later, tremendous efforts have been made to develop one-stage object detector, like SSD \cite{liu2016ssd}, DSSD \cite{fu2017dssd}, and YOLO series \cite{redmon2016you,redmon2017yolo9000,redmon2018yolov3}. RetinaNet \cite{lin2017focal} significantly improves the performance of one-stage detector, making it possible for one-stage detector to rival two-stage detector. Recently, numerous anchor-free approaches have sparked considerable research attention by formulating the objects as key points, such as CornerNet \cite{law2018cornernet}, CenterNet \cite{duan2019centernet}, and FCOS \cite{tian2019fcos}. These methods present a simplified detection scheme,  getting rid of the limitation for anchors. Besides, DETR \cite{carion2020end} pioneers the approach which feeds the serialized features into the Transformer architecture for prediction. DDOD \cite{chen2021disentangle} investigates some conjunctions of training pipeline and proposes a simple yet effective disentanglement method to boost performance.

\subsection{Attention Mechanism}
Recently, inspired by the human perceptual vision system, the attention mechanism has been successfully applied to a wide variety of visual understanding tasks, such as image recognition, image captioning, visual question answering, \etc. The core of attention mechanism is to emphasize the relevant parts while suppressing irrelevant ones. Towards the stable and discriminative representation, considerable developments over it have been investigated in the literature. Specifically, as a pioneering work, RAM \cite{mnih2014recurrent} applies Recurrent Neural Network to locate informative regions in a recursive fashion. Later, SENet \cite{hu2018squeeze} proposes the Squeeze-and-Excitation module to re-calibrate the dependence from channel perspective. Its successors like GSoP-Net \cite{gao2019global} and FcaNet \cite{qin2021fcanet} try to improve the squeeze module. And the follow-up method ECA-Net \cite{Wang_2020_CVPR} pays more attention to the excitation module. The method CBAM \cite{woo2018cbam} jointly explores the inter-channel and inter-spatial relationships between features. Non-Local network [35] captures the long-range dependency and calculates the weighted representations for a certain position in the feature map with the consideration of other positions' contributions. More recently, Gao \etal~\cite{cao2019gcnet} design a lightweight module in conjunction with the simplified Non-Local pipeline. CCNet \cite{huang2019ccnet} presents the Recurrent Criss Cross Attention (RCCA) module which simplifies the global self-attention in Non-Local network.

\subsection{Multi-Label Classification}

Multi-label classification has recently drawn increasing research attention. There are two primary paradigms for multi-label classification, namely, methods based on attention mechanism and those dedicated to exploring the specialized loss functions.

The attention mechanism underpins the first paradigm. CSRA \cite{zhu2021residual} proposes a class-specific residual attention module which adaptively allocates the spatial attention weight to each category. MCAR \cite{gao2021learning} presents an attention mechanism to capture the most informative local region and then feeds the scaled input to the backbone again. Moreover, numerous methods introduce the Transformer \cite{vaswani2017attention} structure. Specifically, Querry2Label \cite{liu2021query2label} introduces a Transformer structure to encode and decode category-related features in the feature map. ML-Decoder~\cite{ridnik2021ml} designs a novel decoder structure  and generalizes to the scenarios with a large number of categories. Another paradigm methods seek to improve performance via the specialized loss functions. For example, Asymmetric Loss \cite{ridnik2021asymmetric} proposes to  dynamically reduce the effect of negative samples and even discard samples suspected to be mislabeled. Class-aware Selective Loss \cite{ben2021multi} proposes a method to estimate the class distribution for the partial labeling problem, and uses a dedicated asymmetric loss in the later training to put more emphasis on the contribution of labeled data rather than unlabeled data.

\subsection{Multi-Scale Feature Fusion}

Multi-scale feature fusion arises naturally in various tasks of computer vision field due to the fact that objects usually show different sizes in the image. Numerous works attempt to utilize the intermediate feature maps of the backbone to improve detection accuracy. For instance, FPN \cite{lin2017feature} pioneeres the approach to fuse features from multiple layers in the CNN-based structure. It establishes a top-down path to provide the informative features to the lower-level feature maps and jointly performs probing at different scales. Later, much effort focuses on how to improve fusion capabilities between different layers in a more efficient manner, \eg, \cite{gong2021effective, pang2019libra, liu2018path, liu2019learning, ghiasi2019fpn, chen2020feature}. Concretely, to enhance the information propagation from upper levels to lower levels, Gong \etal \cite{gong2021effective} propose a fusion factor which play a key role in fusing feature maps at different scales. Liu \etal \cite{liu2019learning} present an adaptive spatial fusion method that greatly improves detection accuracy with little computation overhead. Recently, some more complex  frameworks for feature fusion \cite{ghiasi2019fpn, chen2020feature} have been explored in the literature. NAS-FPN \cite{ghiasi2019fpn} delivers Neural Architecture Search (NAS) to optimize the best path of FPN. Differing from NAS, FPG \cite{chen2020feature} proposes a deep multi-pathway feature pyramid structure to improve the generalization ability.
Unlike the original FPN  structure, CARAFE \cite{wang2019carafe} offers a lightweight learnable module to perform up-sampling. Li \etal \cite{li2022exploring} seek to reconstruct a feature pyramid for the backbones without hierarchical structure.

As many previous works \cite{lin2017feature, liu2018path} show the importance of multi-scale feature fusion, we argue it is the key to solving the problem of prohibited item detection. In X-ray images, many important details of objects are missing, such as texture and appearance information. Moreover, the contours of objects overlap, which also brings great challenges to detection. Multi-scale feature fusion jointly considers the low-level layers with rich detail information and the high-level layers with rich semantic information. Therefore, we propose a dense attention module to capture the relations between feature maps across different stages at inter-channel and inter-pixel positions.

\subsection{Long-tailed Distribution Issue}

The long-tailed distribution issue refers to a small proportion of the categories that accounts for a large proportion of the overall dataset, which makes network struggle to capture discriminative information from the categories in the tail. To this end, tremendous research efforts have been made in the literature, \ie, class re-balancing \cite{kang2019decoupling, wang2019dynamic, cui2019class}, information augmentation \cite{yin2019feature, li2021metasaug}, module improvement \cite{zhou2020bbn, xiang2020learning, cai2021ace}. 

Methods in the first research line focus on the class re-balance strategy. For instance, Kang \etal \cite{kang2019decoupling} decouple the learning process into \emph{representation learning} and \emph{classification} and investigate the influence of different class balancing strategies. DCL \cite{wang2019dynamic} proposes to recognize the tail categories by sampling them more frequently in the later stage of training. Class-Balanced Loss \cite{cui2019class} develops a new re-weighting scheme according to the effective number of samples per class. The methods in the second stream rely on information augmentation. For example, Feature Transfer Learning \cite{yin2019feature} proposes to leverage class information in the head to perform feature amplification on those in the tail. MetaSAug \cite{li2021metasaug} presents an approach based on implicit semantic data augmentation (ISDA) \cite{wang2019implicit} and adopts meta learning to automatically learn to transform semantic orientations. The methods in the last stream draw support from module design perspective.  Bbn \cite{zhou2020bbn} proposes to divide the network into two separate branches, one branch applies naturally randomly and uniformly sample strategy for training, and the other one samples more tail classes. LFME \cite{xiang2020learning} proposes to divide dataset into subsets with smaller “class longtailness” and optimizes multiple models on them individually, and applies knowledge distillation to learn a student model from multiple teacher models.

\section{Dataset PIDray}
\label{dataset}
In this section, we turn to describe about how to construct PIDray dataset in detail, including the data collection, annotation, and statistical information.

\begin{figure}[!t]
	\centering
	\includegraphics[width=\linewidth]{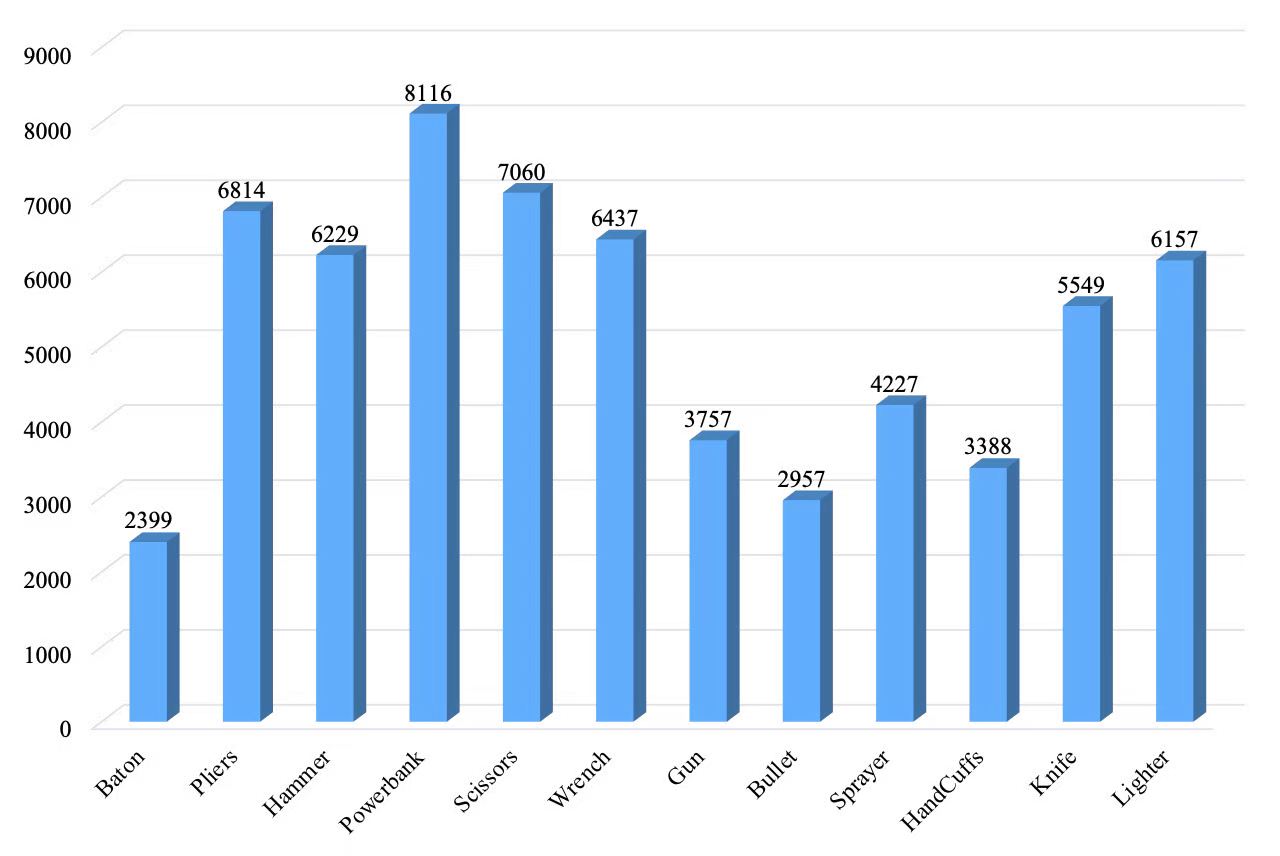}
	\caption{Class distribution of the PIDray dataset. The blue bar represents the number of each class in the PIDray dataset.}
	\label{fig:distribute}
\end{figure}

\subsection{Dataset Collection}
The PIDray dataset is collected in various scenarios (\eg, airports, subway stations, and railway stations), where we are allowed to set up a security inspection machine.  For strong generalization, we deploy $ 3 $ security inspection machines provided by different manufacturers to collect X-ray data. Images generated by different machines usually present certain variances in the size and color of the objects and background. When the packages go through the security inspection machine, they are completely cut out under the guidance of blank parts of the image. In most cases, the X-ray image processes fixed height while its width depends on the size of the package.

The details about the collection process are presented as follows: when the person is required for security inspection, we randomly put the pre-prepared prohibited items in his/her carry-on. Meanwhile, the rough location of the object is recorded, which guarantees that the subsequent annotation work can be carried out smoothly. PIDray dataset is involved with a total of $ 12 $ categories of prohibited items, namely, gun, knife, wrench, pliers, scissors, hammer, handcuffs, baton, sprayer, power bank, lighter and bullet. For the diversity, $ 2 \sim 15 $ instances are ready for each kind of prohibited item. It takes more than six months to collect $ 124, 486$ images for the PIDray dataset. Besides, Fig.~\ref{fig:distribute} summarizes the distribution of tail categories in the dataset. It is notable that all images are saved in the PNG format.
\begin{table}[!t]
	\centering
	\small
	\caption{Statistics of the PIDray dataset.}
	\label{table:partition_dataset}
	\renewcommand{\arraystretch}{1.2}
	\setlength{\tabcolsep}{2.5mm}{
		\begin{tabular}{ccccc}
			\hline
			\multirow{2}[3]{*}{Mode} & \multirow{2}[3]{*}{Train} &   \multicolumn{3}{c}{Test}    \\
			     \cmidrule{3-5}      &                           &   Easy   &  Hard   &  Hidden  \\ \hline
			         Count           &         $76,913$          & $24,758$ & $9,746$ & $13,069$ \\ \hline
			         Total           &               \multicolumn{4}{c}{$124,486$}               \\ \hline
		\end{tabular}}
\end{table}%

\begin{figure}[!t]
	\centering
	\includegraphics[width=1.0 \linewidth]{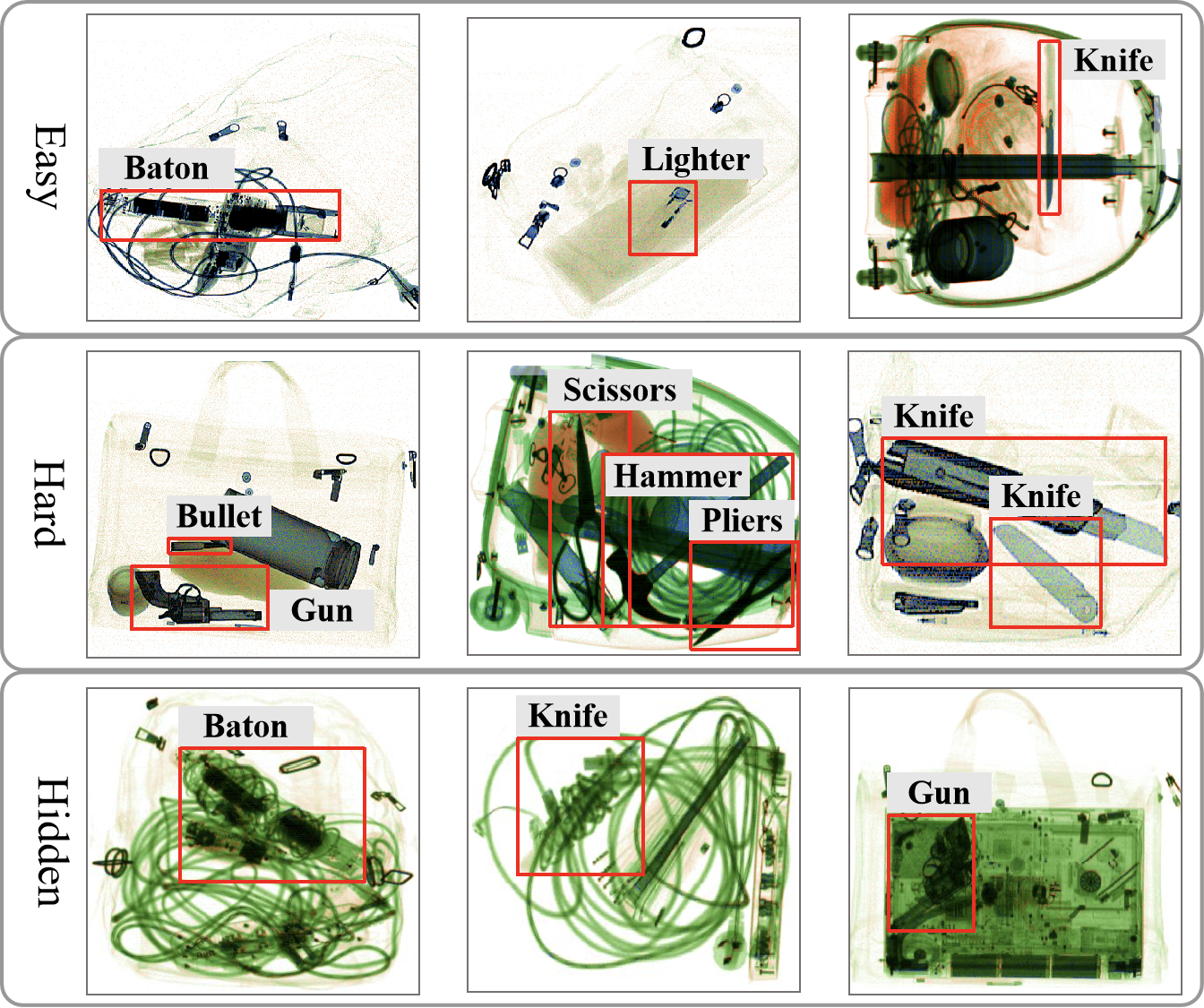}
	\caption{Examples of test sets with different difficulty levels in the proposed PIDray dataset. From top to bottom, the degree of difficulty gradually increases. }
	\label{fig:testset_sample}
\end{figure}

\begin{figure*}[!t]
	\centering
	\includegraphics[width=\linewidth]{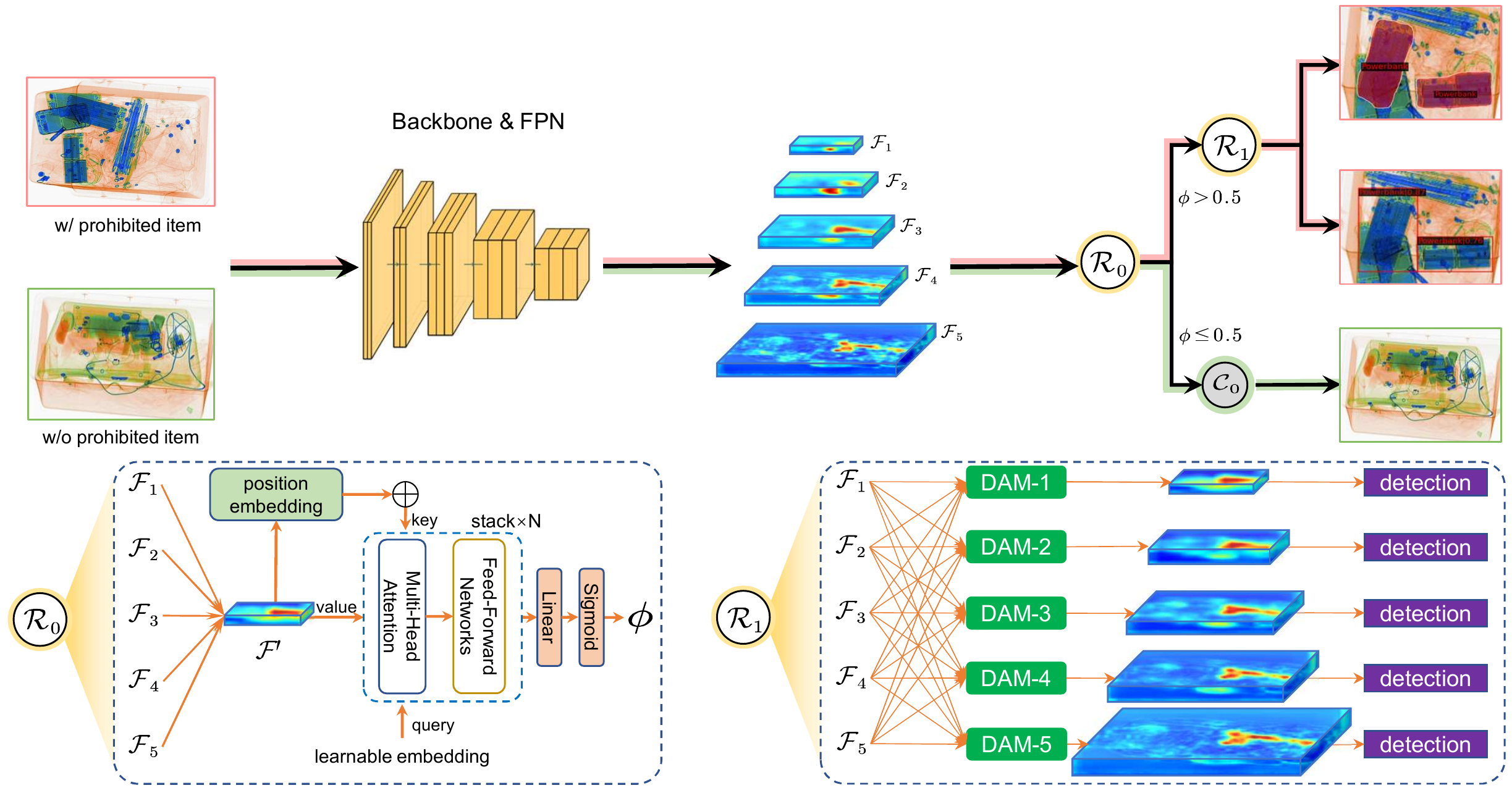}
	\caption{The overall framework of the proposed method for the object detection / instance segmentation task. Specifically, coarse-grained node $ \mathcal{R}_0 $ is tasked with determining whether a sample contains prohibited items or not. The fine-grained node $ \mathcal{R}_1 $ focuses on the task-specific improvements.}
	\label{fig:detection_framework}
\end{figure*}

\subsection{Data Annotation}
Towards high-quality annotations for each image, we offer some training to recruited volunteers in order to identify prohibited items from X-ray images more quickly and accurately. After that, $ 5 $ volunteers are responsible for filtering out samples without any prohibited items from the dataset as well as  annotation of the image-level labels, which greatly facilitate the subsequent annotation work. For the fine-grained annotation, we organize over $ 10 $ volunteers to annotate our dataset with the tool named labelme\footnote{\url{http://labelme.csail.mit.edu/Release3.0/}.}. Each image generally takes about $ 3 $ minutes to annotate, and each volunteer spends about $ 10 $ hours to annotate the image every day. During the annotation process, we label both the bounding box and the segmentation mask of each instance. Next, multiple rounds of double-check are performed to ensure minimum errors.

\subsection{Data Statistics}
To our best knowledge, the PIDray dataset is the largest X-ray prohibited item detection dataset until now. It covers $ 124,486 $ images and $ 12 $ classes of prohibited items. As Table~\ref{table:partition_dataset} shows, we divide those images into $ 76,913 $ (roughly $ 60\% $) images for training and $ 47,573 $ (remaining $ 40\% $) images for testing, respectively. Besides, according to the difficulty degree of prohibited item detection, we split the test set into three subsets, \ie, easy, hard and hidden. In detail, the easy mode means that the image in the test set contains only one prohibited item. The hard mode indicates that the image in the test set contains more than one prohibited item. The hidden mode implies that the image in the test set contains deliberately hidden prohibited items. As shown in Fig.~\ref{fig:testset_sample}, we provide several examples in the test set with different difficulty levels.

\section{Methodology of Object Detection and Instance Segmentation}
\label{object detection}

For object detection and instance segmentation, we adopt the divide-and-conquer pipeline to alleviate the influence of the overwhelming samples without prohibited items on task-specific operations. As Fig.~\ref{fig:detection_framework} illustrates, the input sample first passes through the node $ \mathcal{R}_0 $ to predict whether it contains prohibited items. The node $ \mathcal{R}_0 $ confines a confidence score $ \phi $ in the interval $[0, 1]$. We define that when $ \phi > 0.5 $, it means the current sample contains prohibited items. Otherwise, it means just a normal sample. In our case, only samples with prohibited items need to be fed into the subsequent task-specific processing, which encourages the node $ \mathcal{R}_1 $ to focus more on samples with prohibited items. Meanwhile, we impose the constraint of BCE loss on the output of the node $ \mathcal{R}_0 $ to guide the gradient via back-propagation. Generally, the proportion of samples without prohibited items in the total batch size is uncertain. Hence, we multiply the original loss of the node $ \mathcal{R}_1 $ with the proportion of samples with prohibited items in current batch size to avoid the collapse due to few samples going through the node $\mathcal{R}_1$, which assures the stability of training process.

\subsection{Binary Classification on the Node of $\mathcal{R}_0 $}
Following the principle of divide-and-conquer, the  course-grained node $ \mathcal{R}_0 $  (see the bottom left of Fig.~\ref{fig:detection_framework}) is tasked with extracting features from the multi-scale features $\mathcal{F}_1$-$\mathcal{F}_5$ to perform the binary classification, whose goal is to determine whether the input samples contain prohibited items. To fully unleash the potential of the multi-scale features, we first aggregate these features to obtain a fused feature $ \mathcal{F}^{'} $. Generally, the previous methods usually compress the feature map into a one-dimensional feature via an average pooling operation. Unfortunately, this process tends to leak essential information for class prediction. Instead, we design a cross-attention module, which seeks to facilitate model to encode more informative features. Concretely, we take a learnable embedding as a query, each pixel on the feature map as a value, and each pixel with position embedding as a key. For the query, the global dependencies can be captured by the Equ.~(\ref{eq:1})-(\ref{eq:2}). 

\begin{equation}
	\alpha_i = softmax(\dfrac{W_Q(Q) W_K(E_i)}{\sqrt{d}})
	\label{eq:1}
\end{equation}
\begin{equation}
	\label{eq:2}
	h = \sum_{i=1}^{HW} \alpha_i W_V(F'_i)
\end{equation}
where $W_Q(\cdot)$, $W_K(\cdot)$ and $W_V(\cdot)$ are linear projections for query, key and value respectively, $E$ refers to the spatial embedding obtained from the sum of $\mathcal{F}{'}$ and its position encoding \cite{carion2020end}, $d$ is the dimension of hidden layer, $\alpha_i$ is the score of each position, $h$ denotes the specific head representation. 
Next, we feed the concatenation of the multi-head representation into a feed-forward neural network to obtain the final feature representation, which is expressed as Equ.~(\ref{eq:3}). 
\begin{equation}
	V = FFN(W_H(\prod_{j=1}^{N} h_j))
	\label{eq:3}
\end{equation}
where $\prod$ means the concatenation operation, $W_H(\cdot)$ is the linear projection for fusion of multi-head information, $FFN(\cdot)$ denotes a position-wise feed-forward network \cite{vaswani2017attention}. Finally, we adopt a fully connected layer followed by a sigmoid layer as a classifier to project this feature vector $V$ into a score $\phi$.

\begin{figure*}[ht!]
	\centering
	\includegraphics[width=\linewidth]{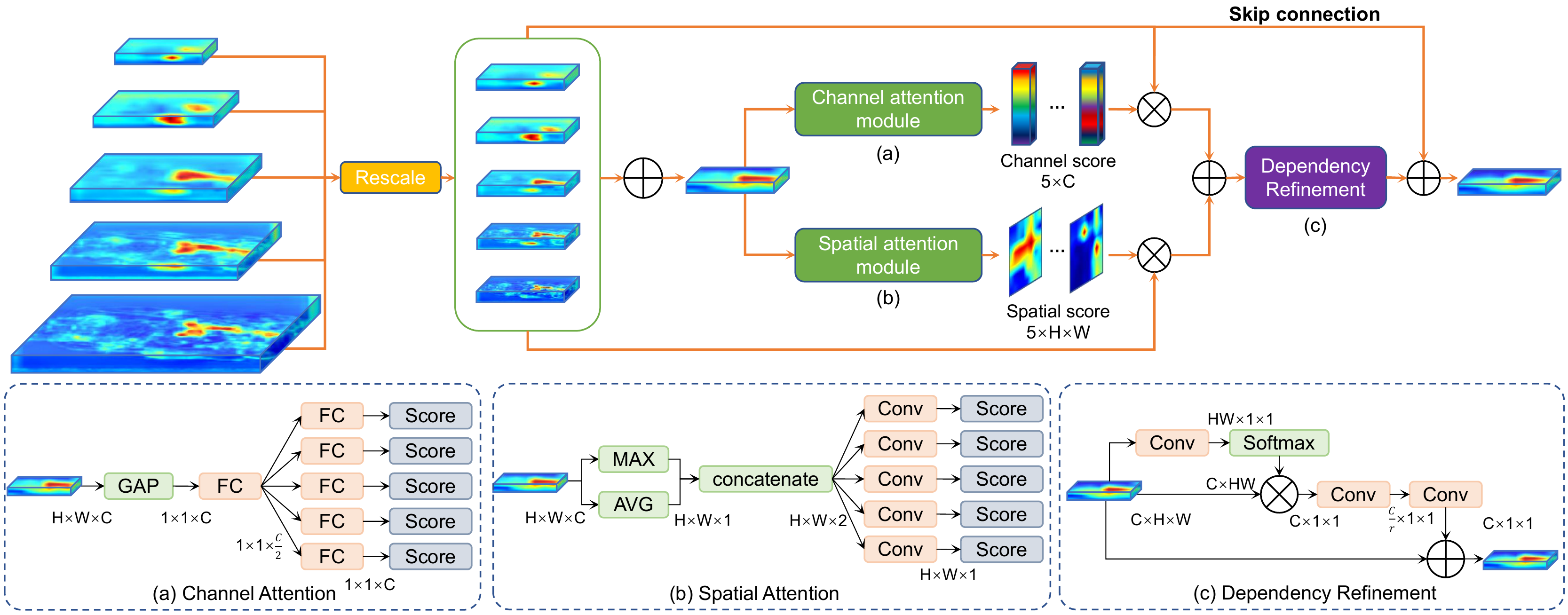}
	\caption{The structure of Dense Attention Modules. DAM is mainly formed by the (a) channel attention, (b) spatial attention and (c) dependency refinement module.}
	\label{fig:DAM-1}
\end{figure*}
\subsection{Detection/Segmentation on the Node of $\mathcal{R}_1 $}
As discussed in the introduction section, the fine-grained node $ \mathcal{R}_1 $ focuses on the task-specific improvements, \ie, contributions to the tasks of object detection and instance segmentation. In our design, we choose the two-stage object detection framework, \eg, Cascade Mask R-CNN \cite{cai2019cascade}, to realize the node $ \mathcal{R}_1 $. Typically, the two-stage object detection frameworks often consist of components such as backbone, Feature Pyramid Network (FPN) \cite{lin2017feature}, detection heads (Region Proposal Network (RPN) followed by the succinct heads). The detector head often works simultaneously on the five feature maps of different scales output by FPN to improve the performance of detection with different sizes. After the RPN provides proposals, the succinct heads (\eg, a simple convolutional layer) are utilized on the pooled feature grid to predict bounding boxes and masks of instances. Nevertheless, the previous approaches usually adopt the feature pyramid structure to exploit multi-scale feature maps in the network, which focus on fusing features only in adjacent layers. The performance tends to be sub-optimal owing to scale variation of objects in complex scenes.

To address the above issue, we design the dense attention module in our study. The proposed framework consists of five Dense Attention Modules (DAMs) (see the bottom right of Fig.~\ref{fig:detection_framework}), which are denoted as $\{DAM_1, DAM_2,\ldots, DAM_5\}$. Each DAM is fed with all multi-scale feature maps in the pyramid as input. Let the feature maps in the pyramid be $\{\mathcal{F}_1,\mathcal{F}_2,\ldots,\mathcal{F}_5\}$. The workflow of the $DAM_i$ is depicted in Fig.~\ref{fig:DAM-1}. Taking the $\mathcal{F}_i$ as an example, we first scale all the input feature maps to match the size of $\mathcal{F}_i$. We adopt max pooling operation for down-sampling feature maps with sizes larger than $\mathcal{F}_i$, while the nearest interpolation for up-sampling those smaller than $\mathcal{F}_i$. Then the five feature maps are added up and fed into the Channel-wise Attention Module and Spatial-wise Attention Module to obtain channel-wise and spatial-wise weights respectively. In another word, we recalibrate the importance of various feature maps from the channel and space perspectives. Subsequently, multiplication operation is utilized to emphasize the attentive channel and spatial regions. After that, the sum of the channel and spatial features is fed into the dependency refinement module. Finally, we add this result directly to $\mathcal{F}_i$ via a lateral skip connection. 
Note that above three steps are performed on the feature map in each layer. After assembling both the original and enhanced maps, the multi-scale representation is fed into the detection heads for final prediction.

\textbf{Channel-wise Attention Module.} As shown in Fig.~\ref{fig:DAM-1} (a), we feed the channel-wise attention module with the aggregated representation to allocate the channel-wise weight to various feature maps. In detail, the aggregated representation is squeezed along the channel-wise dimension by a fully connected layer. The derived vector is then fed into the fully connected layers followed by the softmax layers to obtain the corresponding channel scores.

\textbf{Spatial-wise Attention Module.} As Fig.~\ref{fig:DAM-1} (b) illustrates, the aggregated representation is fed into the spatial-wise attention module to recalibrate the spatial-wise weight of the feature maps. Unlike the channel-wise attention module, the pooling operation here is performed along the channel dimension. We first impose max pooling and average pooling on the input individually. Then we concatenate these two feature maps and feed them into five convolutional layers followed by softmax layers to obtain the spatial-wise scores of $H\times W$. 

\begin{figure*}[h!]
	\centering
	\includegraphics[width=0.935\linewidth]{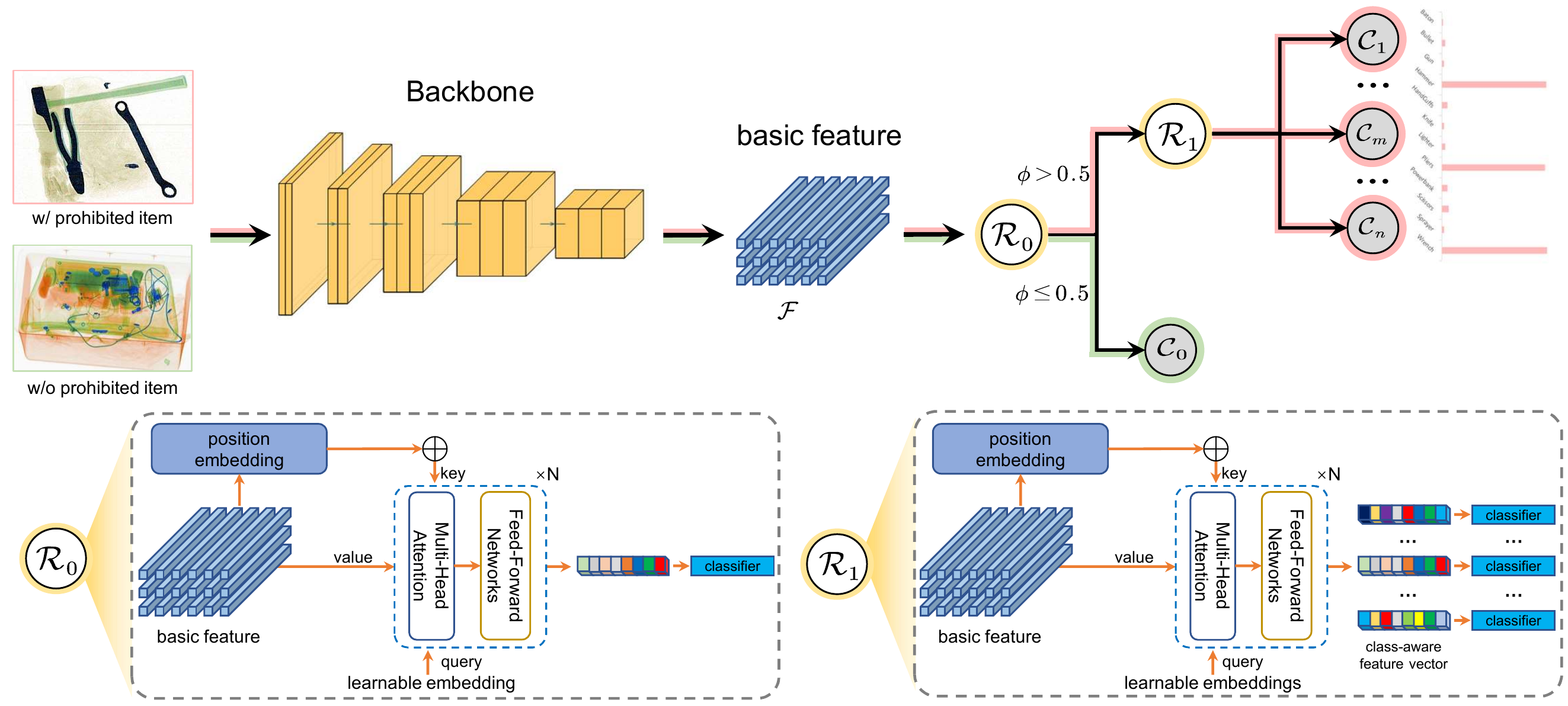}
	\caption{The overall architecture of the proposed method for the multi-label classification task. The course-grained node $ \mathcal{R}_0 $ is tasked with a binary classification while the fine-grained node $\mathcal{R}_1  $ is dedicated to the multi-label classification of the categories of prohibited items.}
	\label{fig:multilabel_network}
\end{figure*}

\textbf{Dependency Refinement.} The spatial and channel attention adopted earlier primarily focus on the fusion of the five feature maps according to their contribution. Generally, it plays an indispensable role for the better performance to construct the long-range dependency between distant pixels. Non-Local Block \cite{wang2018non} is featured of efficient acquisition of long-range dependency. Thus, we introduce its simplified version to compensate for this drawback in the proposed method. As Fig.~\ref{fig:DAM-1} (c) depicts, towards the accurate results, we introduce such a module to perform dependency refinement for the aggregated feature maps.

\section{Methodology of Multi-Label Classification}
\label{multi-label classification}
Since the proposed dataset can serve as a benchmark dataset for the multi-label classification task as well, we extend the spirit of divide-and-conquer to this task. In specific, we first feed the input image $\mathcal{I}$ into the backbone to obtain the basic feature map $\mathcal{F}$. Previous methods usually feed this representation directly into the multi-label classifier to finalize the prediction. However, this scheme is subject to the long-tailed issue heavily, resulting in the degraded performance. Based on the observation of the distribution in our proposed dataset,  we propose a tree-like framework for this task. As in Fig.~\ref{fig:multilabel_network}, in our design, the course-grained node is responsible for the prediction to determine whether prohibited items exist, while the fine-grained node focuses on predicting the specific prohibited classes. Interestingly, such a simple yet efficient scheme enables us to enjoy significant performance over the state-of-the-art methods.

\subsection{Binary Classification on the Node $ \mathcal{R}_0 $}
As stated above, the course-grained node $ \mathcal{R}_0 $ is tasked with the binary classification based on the basic feature map $ \mathcal{F} $. In our design, we realize the node $ \mathcal{R}_0 $ with a light-weight network which is mainly composed of a multi-head attention module \cite{vaswani2017attention}, a position-wise feed-forward network (FFN) and a binary classifier, see the structure in the bottom left of Fig.~\ref{fig:multilabel_network}. In detail, we treat a learnable embedding as the query, the feature map with position embedding as the key, and the original feature map as the value. Following the Equ.~(\ref{eq:1})-(\ref{eq:3}), we explore the global dependencies for the query in terms of specific head, then we feed the concatenation of multi-head representation into a feed-forward neural network to generate the final feature representation. Experimentally, we observe that this scheme enables our model to pay more attention to the discriminative regions. Finally, the feature representation is projected to a score $\phi$ with a binary classifier.

\subsection{Multi-label Classification on the Node $\mathcal{R}_1 $}
With the help of node $ \mathcal{R}_0 $, the fine-grained node $ \mathcal{R}_1 $ is dedicated to the multi-label classification of the categories of prohibited items. For convenience, we reuse the structure design of the node $ \mathcal{R}_0 $. The node $ \mathcal{R}_1 $ is responsible for class-aware representation according to the class-specific query. This module allocates different attention weight in the space of the feature map according to specific queries and generates the class-aware representations. Finally, the class-aware representations are mapped into a score via an independent binary classifier. The only difference is the node $ \mathcal{R}_0 $ needs a binary classifier while the node $ \mathcal{R}_1 $ is equipped with multiple binary classifiers. The structure $ \mathcal{R}_1 $ in the bottom right of Fig.~\ref{fig:multilabel_network} illustrates the details.

\subsection{Loss Function}
Our loss function mainly consists of two parts. One is the binary cross-entropy loss of the node $ \mathcal{R}_0 $, which is formulated as Equ.~(\ref{eq:BCELoss}).

\begin{equation}
	\mathcal{L}_{bc}=-[y\ln(p) + (1-y)\ln(1-p)]
	\label{eq:BCELoss}
\end{equation}
where $p$ is the prediction confidence, and $ y $ is the corresponding binary label.

The other one is the multi-label classification of the node $ \mathcal{R}_1 $, here we adopt the asymmetric loss \cite{ridnik2021asymmetric}, which is presented as Equ.~(\ref{eq:ASLLoss}).

\begin{equation}
	\mathcal{L}_{ml}=\frac{1}{C} \sum_{k=1}^{C}\left\{\begin{array}{ll}
		\left(1-p_{k}\right)^{\gamma+} \ln \left(p_{k}\right), & y_{k}=1 \\
		\left(p_{k}\right)^{\gamma-} \ln \left(1-p_{k}\right), & y_{k}=0
	\end{array}\right.
	\label{eq:ASLLoss}
\end{equation}
where $p_k$ is the prediction of the $k$-th class, and $y_k$ is the corresponding label of the $k$-th class, $\gamma+$ and $\gamma-$ are two hyper-parameters.

In a nutshell, the total loss function is expressed as below:
\begin{equation}
	\mathcal{L} = \lambda *\mathcal{L}_{ml} +  \mathcal{L}_{bc}
\end{equation}
where $ \lambda $ is the hyper-parameter to balance two loss items, its value in our experiments is set to be the proportion of samples with prohibited items in batch size.

\section{Experiments}
\label{experiments}

In this section, we conduct extensive experiments on the PIDray dataset to validate effectiveness of the proposed method systematically. To be specific, we first describe implementation details, evaluation metrics, overall evaluation and ablation studies. Then the performance comparisons between the proposed method and the state-of-the-art methods on the PIDray dataset are reported. Next, the experiments on general image datasets (\ie, COCO and PASCAL VOC)  are performed to demonstrate the generalization ability of the proposed model. Finally, we present the insightful analysis to verify the importance of key components or hyper-parameters via ablation studies.

\begin{table*}[t!]
	\caption{Overall evaluation of object detection and instance segmentation.}
	\label{tab:detection overall evaluation}
	\begin{center}
		\renewcommand{\arraystretch}{1}
		\setlength{\tabcolsep}{1.3mm}
		\begin{tabular}{@{}cccccccccc@{}}
			\hline
			              \multirow{2}{*}{Method}                & \multirow{2}{*}{Backbone} &                   \multicolumn{4}{c}{Detection AP}                   &                  \multicolumn{4}{c}{Segmentation AP}                  \\ 
			                         \cmidrule(l){3-6} \cmidrule(l){7-10}                           &                           &      \rotatebox{90}{Easy}       &      \rotatebox{90}{Hard}       &     \rotatebox{90}{Hidden}      &     \rotatebox{90}{Overall}     &      \rotatebox{90}{Easy}       &      \rotatebox{90}{Hard}       &     \rotatebox{90}{Hidden}      &     \rotatebox{90}{Overall}     \\ \hline
			              FCOS \cite{tian2019fcos}               &      ResNet-101-FPN       &      57.9       &      47.4       &      28.5       &      48.2       &        -        &        -        &        -        &        -        \\
			           RetinaNet \cite{lin2017focal}             &      ResNet-101-FPN       &      65.8       &      55.1       &      43.2       &      57.6       &        -        &        -        &        -        &        -        \\
			              SSD512 \cite{liu2016ssd}               &           VGG16           &      65.9       &      57.6       &      44.7       &      58.9       &        -        &        -        &        -        &        -        \\
			              TOOD \cite{feng2021tood}               &      ResNet-101-FPN       &      68.2       &      63.2       &      45.3       &      62.2       &        -        &        -        &        -        &        -        \\
			                DW \cite{li2022dual}                 &      ResNet-101-FPN       &      67.9       &      61.3       &      46.8       &      61.6       &        -        &        -        &        -        &        -        \\
			          DDOD \cite{chen2021disentangle}            &      ResNet-101-FPN       &      69.5       &      63.6       &      48.9       &      63.6       &        -        &        -        &        -        &        -        \\\hline
			                     DDOD+Ours                       &      ResNet-101-FPN       & \red{\bf{71.2}} & \red{\bf{64.3}} & \red{\bf{50.2}} & \red{\bf{64.7}} &        -        &        -        &        -        &        -        \\ \hline
			         Faster R-CNN \cite{ren2015faster}           &      ResNet-101-FPN       &      65.8       &      56.9       &      44.2       &      58.3       &        -        &        -        &        -        &        -        \\
			          Libra R-CNN \cite{pang2019libra}           &      ResNet-101-FPN       &      64.0       &      55.6       &      42.1       &      56.7       &        -        &        -        &        -        &        -        \\
			            Mask R-CNN \cite{he2017mask}             &      ResNet-101-FPN       &      66.2       &      58.6       &      43.8       &      59.1       &      59.2       &      50.1       &      35.5       &      51.2       \\
			        Cascade R-CNN \cite{cai2019cascade}          &      ResNet-101-FPN       &      70.5       &      61.2       &      49.1       &      63.1       &        -        &        -        &        -        &        -        \\
			      Cascade Mask R-CNN \cite{cai2019cascade}       &      ResNet-101-FPN       &      71.9       &      63.2       &      46.8       &      63.7       &      60.7       &      52.0       &      36.2       &      52.8       \\
			Cascade Mask R-CNN \cite{cai2019cascade,liu2018path} &     ResNet-101-PAFPN      &      72.2       &      63.7       &      48.3       &      64.4       &      61.1       &      52.0       &      37.0       &      53.1       \\\hline
			           SDANet  \cite{wang2021towards} (our conference)             &      ResNet-101-FPN       &      72.5       &      63.7       &      48.0       &      64.4       &      61.1       &      51.7       &      37.0       &      52.9       \\ 
			              Cascade Mask R-CNN+Ours                &      ResNet-101-FPN       & \red{\bf{74.5}} & \red{\bf{64.8}} & \red{\bf{53.0}} & \red{\bf{66.6}} & \red{\bf{61.4}} & \red{\bf{51.9}} & \red{\bf{39.7}} & \red{\bf{53.4}} \\ \hline
		\end{tabular}
	\end{center}
\end{table*}
\subsection{Implementation Details}
We adopt the MMDetection toolkit as our training platform, which is performed on a machine with four NVIDIA RTX 3090 GPUs. Our method is implemented in PyTorch deep learning framework. For a fair comparison, all the compared methods are trained on the training set and evaluated on the test set of the PIDray dataset. In terms of object detection and instance segmentation tasks, the proposed pipeline is realized on top of Cascade Mask-RCNN [5], where the ResNet-101 network is used as the backbone. According to our statistics, the average resolution of the images in our dataset is about $ 500 \times 500 $. Hence, we resize the image to $ 500\times500 $ for compared detectors. The entire network is optimized with a stochastic gradient descent (SGD) algorithm with a momentum of $ 0.9 $ and a weight decay of $ 0.0001 $. The initial learning rate is set $ 0.02 $ and the batch size is set $ 16 $. We train $ 12 $ epochs and  $ 24 $ epochs for two-stage and the one-stage detectors, respectively. Unless otherwise specified, other hyper-parameters involved in the experiments follow the settings of MMDetection. 

For multi-label classification task, we perform all experiments on a machine with eight NVIDIA RTX 3090 GPUs. The proposed model is trained for $ 80 $ epochs with an early stopping strategy. We use the Adam optimizer with True-weight-decay \cite{loshchilov2017decoupled} of $1\times10^{-2}$ and the one cycle policy \cite{smith2019super} to optimize the proposed model. Notably, for ResNet-101 based models, we set the maximum learning rate of $8\times10^{-5}$ and batch size of 288. For TResNetL based models, we set the maximum learning rate of $1.2\times10^{-4}$ and batch size of 288. For CvT-21-384 based models, we set the maximum learning rate of $1\times10^{-4}$ and batch size of 240. For CvT-w24-384 we set the maximum learning rate of $5\times10^{-5}$ and batch size of 40. In terms of hyper-parameters $\gamma^+$ and $\gamma^-$ in the Equ.~(\ref{eq:ASLLoss}), we set $ 0 $ and $ 2 $ for ResNet-101 based models and $  0 $ for other models.
\subsection{Evaluation Metrics}
Following common metrics of MS COCO \cite{lin2014microsoft}, we evaluate the performance of the compared methods in terms of the AP and AR metrics on our PIDray dataset. The scores are averaged over multiple Intersection over Union (IoU). Notably, we use $ 10 $ IoU thresholds between $ 0.50 $ and $ 0.95 $. Specifically, the AP score is averaged across all $ 10 $ IoU thresholds and all $ 12 $ categories. In order to better assess a model, we look at various data splits. $ AP_{50} $ and $ AP_{75} $ scores are calculated at IoU = $  0.50 $ and IoU = $ 0.75 $ respectively. Note that many prohibited items are small (area $  < 32^2 $) in the PIDray dataset, which is evaluated by the $ AR_S $ metric. Besides, the AR score is the maximum recall given a fixed number of detections (\eg, $ 1 $, $ 10 $, $ 100 $) per image, averaged over $ 12 $ front categories and $ 10 $ IoUs.

In terms of the multi-label classification task, we use mean Average Precision (mAP) as an evaluation metric. It first calculates the Average Precision(AP) for each category, \ie, the area of the Precision-Recall curve, and then averages the AP over all categories.

\subsection{Overall Evaluation}
\begin{figure*}[ht]
	\centering
	\includegraphics[width=1.0\linewidth]{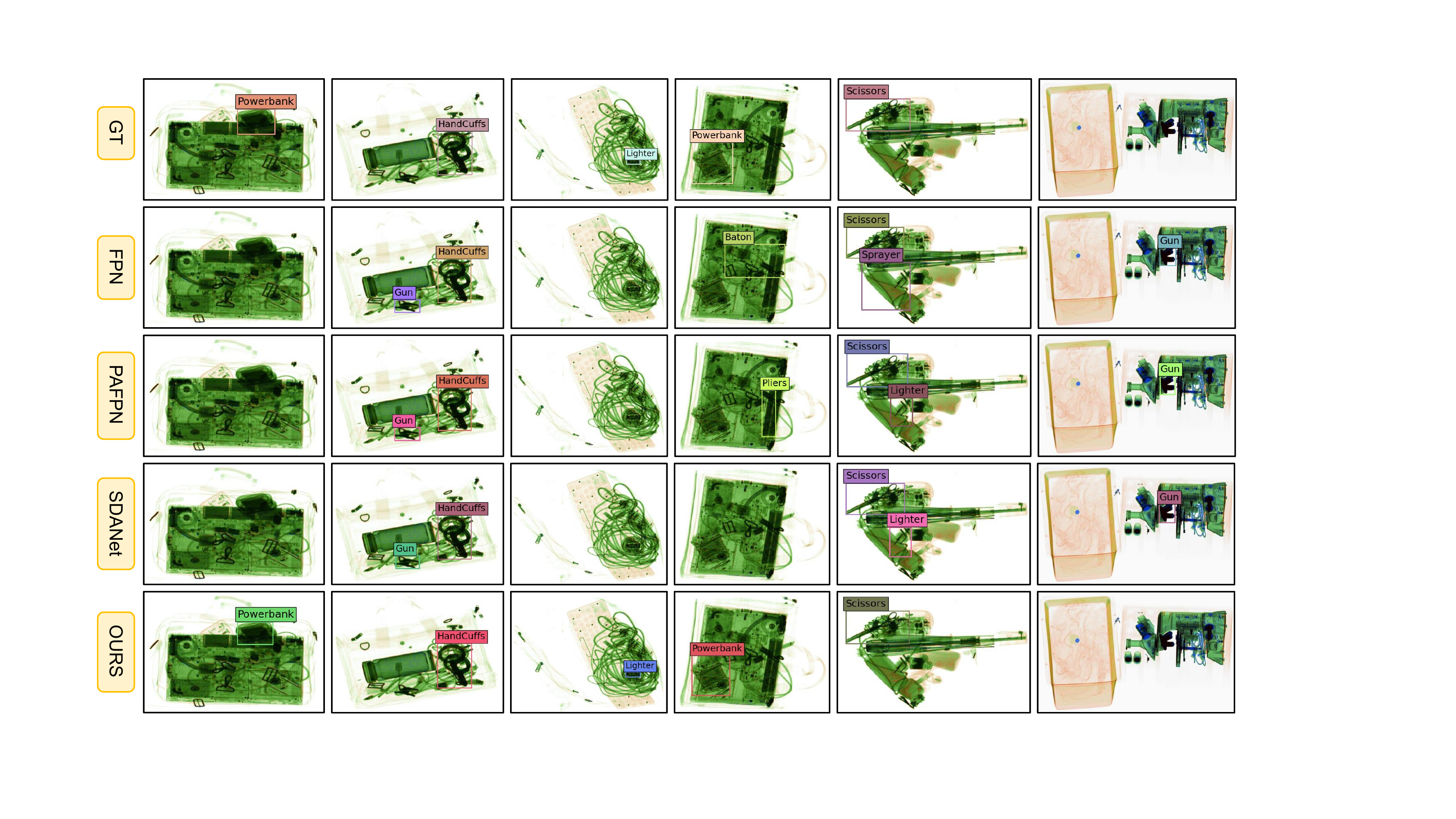}
	\caption{Visual comparison on detection task. GT indicates Ground-truth, FPN denotes the results generated by Cascade Mask R-CNN with FPN, PAFPN denotes the results generated by Cascade Mask R-CNN with PAFPN, SDANet denotes the results generated by SDANet and Ours indicates the results generated by the proposed method.}
	\label{figure:bbox_comparison}
\end{figure*}

\begin{figure*}[h!]
	\centering
	\includegraphics[width=\linewidth]{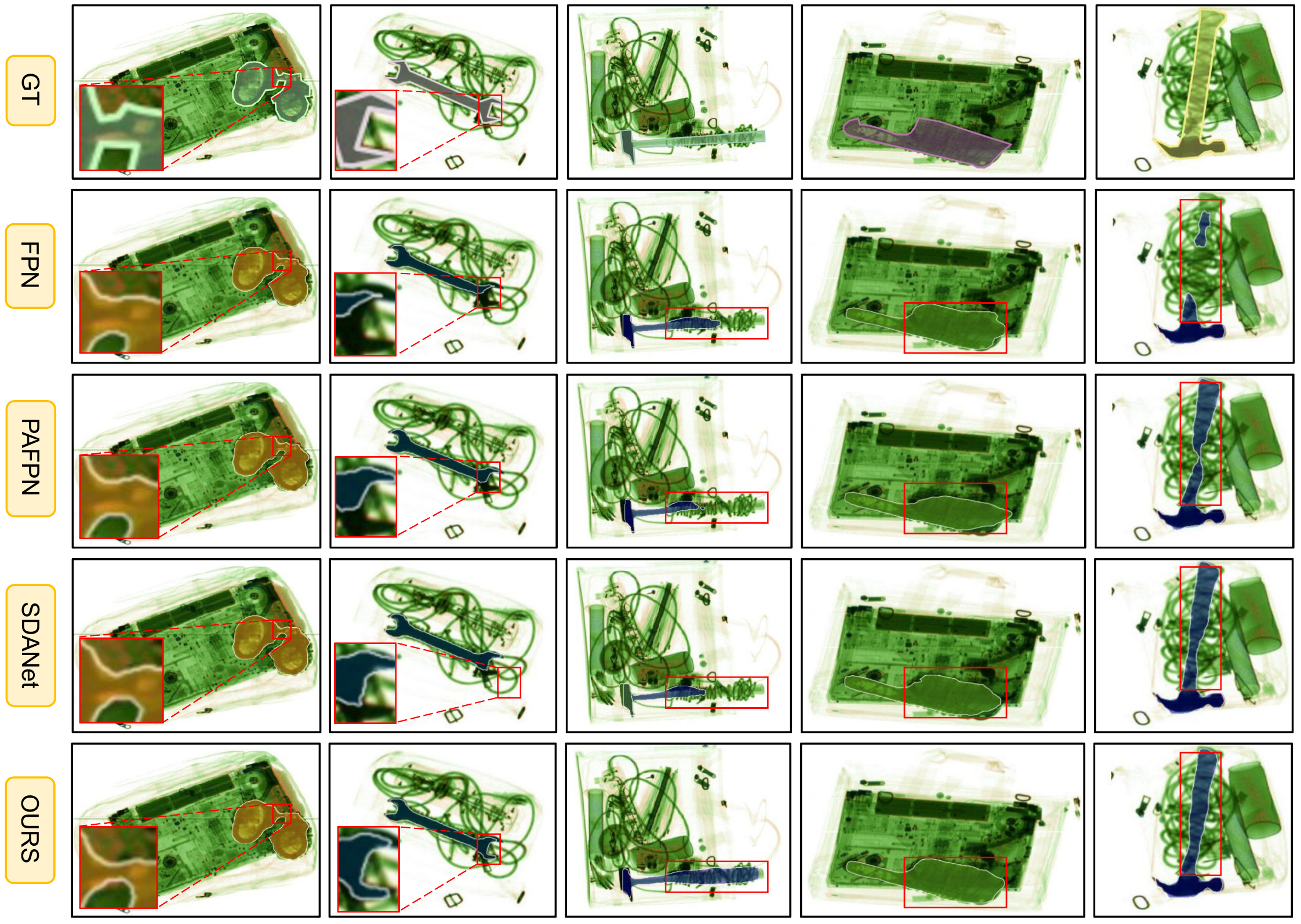}
	\caption{Visual comparison on instance segmentation task. GT indicates Ground-truth, FPN denotes the results generated by Cascade Mask R-CNN with FPN, PAFPN denotes the results generated by Cascade Mask R-CNN with PAFPN, SDANet denotes the results generated by SDANet and Ours indicates the results generated by the proposed method.}
	\label{figure:seg_comparison}
\end{figure*}

Firstly, as presented in Table \ref{tab:detection overall evaluation}, we report quantitative performance comparisons between our method and numerous one-stage or two-stage state-of-the-art object detectors. Notably, since some methods do not support the instance segmentation task in literature, we fill the placeholder `-' in Table \ref{tab:detection overall evaluation} for illustration. One can see that our method achieves the superior performance in terms of all metrics on various subsets of the PIDray dataset. For example, compared with the second-best performed methods, our method gains absolute $3.9\%$ and $2.7\%$ AP gain for the two sub-tasks on the hidden test set, which strongly demonstrates effectiveness of the proposed pipeline. As evidenced in Fig.~\ref{figure:bbox_comparison} and Fig.~\ref{figure:seg_comparison}, our method shows obvious advantages over other methods. For example, in the first column of Fig.~\ref{figure:bbox_comparison}, all methods except ours miss the powerbank which is deliberately hidden in the messy objects. In the second and fifth columns, one can see that these baselines over-identify or mis-identify some items. In the sixth column, these baselines are prone to error when processing the image without any prohibited items.  The Fig.~\ref{figure:seg_comparison} exhibits a similar visual trend. For instance, in the second, third and fifth columns, visual results show that these baselines tend to generate the incomplete masks in the boxed area. In the fourth column, incomplete coverage or over-flowing masks characterize the results generated by baselines. These visual comparisons reveal that the previous baselines are challenged to capture the features of hidden items, while our approach detects prohibited items effectively, especially those that have been deliberately hidden. We believe it is because the proposed pipeline endows model with a more comprehensive understanding of the characteristics of the proposed dataset.
Moreover, our contributions to the task-specific node can bring the following two benefits. First, attention-wise modules can propagate semantic information across multi-layers densely. Second, the dependency refinement module can explore long-range dependencies among different feature maps. These complementary design choices make our method can detect deliberately hidden data effectively.

\begin{table}[!t]
	\caption{Overall evaluation of Multi-labeling classification.}
	\label{tab:SOTA}
	\begin{center}
		\renewcommand{\arraystretch}{1.2}
		\setlength{\tabcolsep}{.45mm}
		\begin{tabular}{@{}c | c | c | c@{}}
			\hline
			            Method              & Backbone   & Resolution & mAP                                  \\ \hline
			  CSRA \cite{zhu2021residual}   & ResNet-101 & 448x448    & 86.94                                \\
			 Q2L \cite{liu2021query2label}  & ResNet-101 & 448x448    & 87.69                                \\
			             Ours               & ResNet-101 & 448x448    & \red{\bf{89.22}} \blue{\bf{(+1.53)}} \\ \hline
			ASL \cite{ridnik2021asymmetric} & TResNetL   & 448x448    & 92.91                                \\
			 Q2L \cite{liu2021query2label}  & TResNetL   & 448x448    & 92.99                                \\
			ML-Decoder \cite{ridnik2021ml}  & TResNetL   & 448x448    & 92.81                                \\
			    CSL \cite{ben2021multi}     & TResNetL   & 448x448    & 93.15                                \\
			             Ours               & TResNetL   & 448x448    & \red{\bf{93.44}} \blue{\bf{(+0.29)}} \\ \hline
			 Q2L \cite{liu2021query2label}  & CvT-21     & 384x384    & 91.71                                \\
			             Ours               & CvT-21     & 384x384    & \red{\bf{93.16}} \blue{\bf{(+1.45)}} \\ \hline
			 Q2L \cite{liu2021query2label}  & CvT-w24    & 384x384    & 93.49                                \\
			             Ours               & CvT-w24    & 384x384    & \red{\bf{94.24}} \blue{\bf{(+0.75)}} \\ \hline
		\end{tabular}
	\end{center}
\end{table}

Secondly, we summarize the experimental performance between our method and the state-of-the-art works on the PIDray dataset for multi-labeling classification task in Table \ref{tab:SOTA}. Compare with these approaches, the proposed method shows the obvious superior performance, regardless of the backbone or image resolution. In detail, for approaches based on ResNet-101 \cite{he2016deep}, our method achieves the best result, outperforming the second-best method Q2L \cite{liu2021query2label} by an absolute improvement of $ 1.53 $ point in terms of the mAP. With regard to the methods built on top of TResNetL \cite{ridnik2020tresnet}, the proposed method shows superiority over them as well. When couple with the CvT \cite{wu2021cvt} series backbone, one can see that our method performs best consistently, boosting accuracy by $ 1.45 $ and $ 0.75 $ respectively. It can be concluded that these state-of-the-art methods still face challenges due to the obvious discrepancies between X-ray and natural images. While our method relies on the divide-and-conquer pipeline and the attention mechanism to alleviate this issue in an effective manner.

\begin{table}[!t]
	\caption{Evaluation results on the MS COCO and PASCAL VOC detection datasets.}
	\centering
	\renewcommand{\arraystretch}{1.2}
	\setlength{\tabcolsep}{0.5mm}
	\begin{tabular}{@{}c|c|c|c@{}}
		\hline
		 Method   & MS COCO2014 & MS COCO2017 & PASCAL VOC \\ \hline
		Baseline &    40.8     &    42.8     &    81.7    \\ \hline
		  Ours    &  {\bf41.6}  &  {\bf43.2}  & {\bf82.1}  \\ \hline
	\end{tabular}%
	\label{tab:general_datasets}%
	\vspace{-0.2cm}
\end{table}%

Finally, we shift attention towards general detection datasets to validate the generalization ability of the proposed method on the natural image. The experiments are performed on  MS COCO \cite{lin2014microsoft} and PASCAL VOC \cite{everingham2010pascal}, which are quite popular datasets for natural image detection domain. For a fair comparison, we follow the default experiment settings in MMDetection. The experimental results are reported in Table \ref{tab:general_datasets}.  Compared with the baseline methods (Cascade Mask R-CNN for MS COCO2014 and MS COCO2017, Cascade R-CNN for PASCAL VOC), we have achieved $ +0.8 $ AP,  $ +0.4 $ AP and $ +0.4$ AP gain on MS COCO2014, MS COCO2017 and PASCAL VOC, respectively. Experimental results demonstrate that our method is not only effective for the detection of prohibited items, but also suitable for general scenarios.

\subsection{Ablation Study}
In this section, we perform ablation studies on the necessity of samples without prohibited items, effect of divide-and-conquer pipeline, contributions of the fine-grained node to specific tasks.

\subsubsection{Necessity of samples without prohibited items}

\begin{figure}[!t]
	\centering
	\includegraphics[width=\linewidth]{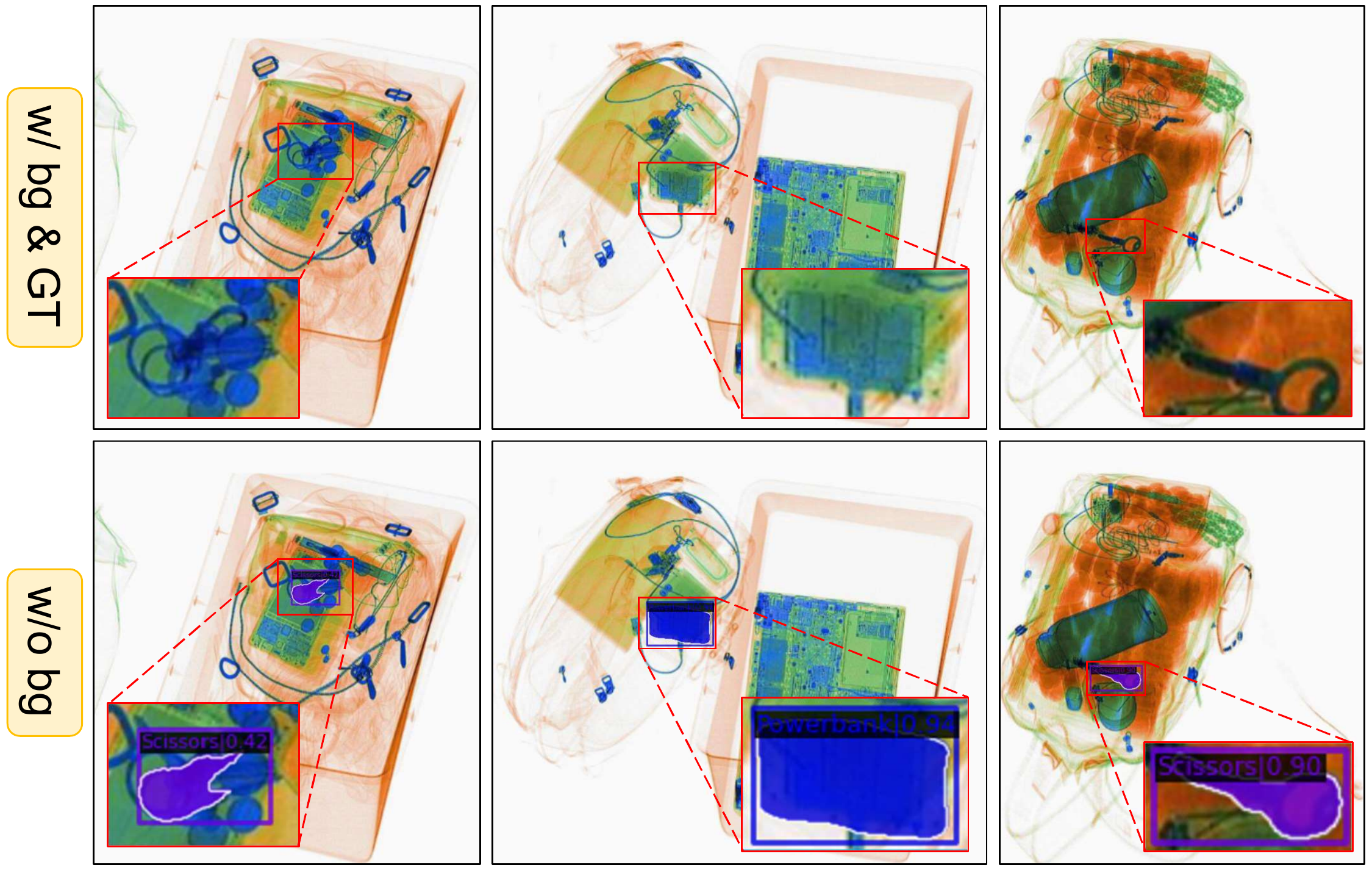}
	\caption{The influence of images without prohibited items.}
	\label{fig:train_set_comparison}
\end{figure}

In real-world scenarios, it is not always true that only images with prohibited items need to be detected. Hence, an open question is if a model pre-trained on the dataset of which samples with target object account for the majority can work well in security inspection. As Table \ref{tab:detection influence of dataset} shows, we report the necessity of samples without prohibited items in the PIDray dataset for object detection task in terms of the mAP and Error Rate. We adopt the Cascade Mask R-CNN \cite{cai2019cascade} as the test model. During the test, we define the image of which any bounding box with a confidence level greater than $ 0.5 $ to be a sample with prohibited items. By comparison, we can conclude that the introduction of samples without prohibited items to the training set not only improves detection accuracy $ (+0.5) $, but also significantly reduces the error rate ($-16.6 $). And as evidenced in Fig.~\ref{fig:train_set_comparison}, a model trained on a training set excluding samples that do not contain prohibited items still has an item circled on the normal test image.

We also verify this argument on the multi-label classification task. The neural model should predict $0$ for each category when no prohibited items are present. We use the Q2L with CvT-w24-384 backbone as the test model. In terms of Error Rate, we define the input as a sample without prohibited items when all twelve categories in the dataset present a confidence score below $0.5$. As shown in Table \ref{tab:multi-label influence of dataset}, performance on the full dataset significantly improves mAP (+3.45) and reduces the error rate (-61.7). Further, we report the rate of false positive rate (FP). It can be clearly seen that if there are no samples without prohibited items in the training dataset, the network tends to guess wildly at least one category, leading to an oddly high rate of misjudgments.

Based on the above observation, we conclude that there is a strong necessity to introduce samples without prohibited items into the PIDray dataset.

\begin{table}[!t]
	\caption{Influence of samples without prohibited items. Error Rate denotes determining whether or not they contain prohibited items}
	\label{tab:detection influence of dataset}
	\begin{center}
		\renewcommand{\arraystretch}{1.2}
		\setlength{\tabcolsep}{1mm}
		\begin{tabular}{@{}c | c | c@{} }
			\hline
			     Training Dataset      & Detection AP & Error Rate \\ \hline
			w/o Non-Prohibited Samples & 63.2         & 21.4       \\
			       Full Dataset        & 63.7         & 4.8        \\ \hline
		\end{tabular}
	\end{center}
\end{table}

\subsubsection{Effect of the divide-and-conquer pipeline}

To verify the effect of our proposed divide-and-conquer pipeline, we choose Cascade Mask R-CNN \cite{cai2019cascade} and DDOD \cite{chen2021disentangle} as the baseline to add a coarse-grained node between backbone and detection head. As Table \ref{tab:ablation_coarse_node} shows, we achieve significant improvements (+0.9 and +1.7 on detection AP) over the two baselines. We believe it is because the original fine-grained node can focus more on the samples with prohibited items with the help of coarse-grained node. When detached from the coarse-grained node, the task-specific node is challenged by overwhelming samples without prohibited items, resulting in degraded performance.

\subsubsection{contributions of fine-grained node to object detection/ instance segmentation}

\begin{table}[!t]
	\caption{Influence of samples without prohibited items}
	\label{tab:multi-label influence of dataset}
	\begin{center}
		\renewcommand{\arraystretch}{1.2}
		\setlength{\tabcolsep}{1mm}
		\begin{tabular}{@{}c | c | c | c@{}}
			\hline
			      Train Dataset        & mAP   & \tabincell{c}{Error \\ Rate}         & \tabincell{c}{FP \\Rate}       \\ \hline
			w/o Non-Prohibited Samples & 90.04 & 61.70              & $\approx $100 \\
			       Full Dataset        & 93.49 & $6.3\times10^{-3}$ & $\approx $0   \\ \hline
		\end{tabular}
	\end{center}
\end{table}

\begin{table*}[ht]\scriptsize
	\caption{ Effectiveness of divide-and-conquer pipeline. All models are trained on the PIDray {\it training} subset and tested on the PIDray {\it test} set. The accuracies are denoted by ``detection AP/segmentation AP''. }
	\label{tab:ablation_coarse_node}
	\begin{center}
		\renewcommand{\arraystretch}{1.2}
		\setlength{\tabcolsep}{1mm}
		\begin{tabular}{c|cccccccc}
			\hline
			      Method       &     AP      & $\text{AP}_{50}$ & $\text{AP}_{75}$ & $\text{AP}_{S}$ & $\text{AR}_{1}$ & $\text{AR}_{10}$ & $\text{AR}_{100}$ & $\text{AR}_{S}$ \\ \hline
			       DDOD        &  $63.6/-$   &     $79.0/-$     &     $70.1/-$     &    $71.2/-$     &    $66.4/-$     &     $71.5/-$     &     $71.6/-$      &    $71.6/-$     \\
			    w/ Pipline     &  $64.5/-$   &     $79.9/-$     &     $71.2/-$     &    $71.7/-$     &    $67.0/-$     &     $72.1/-$     &     $72.2/-$      &    $72.2/-$     \\ \hline
			Cascade Mask R-CNN & $63.7/52.5$ &   $78.6/76.3$    &   $71.1/60.6$    &   $70.1/57.7$   &   $66.7/55.9$   &   $70.4/58.3$    &    $70.4/58.3$    &   $70.4/58.3$   \\
			    w/ Pipline     & $65.4/52.7$ &   $80.3/77.6$    &   $72.8/60.5$    &   $71.4/57.5$   &   $68.0/55.9$   &   $71.7/58.2$    &    $71.7/58.2$    &   $71.7/58.2$   \\ \hline
		\end{tabular}
	\end{center}
\end{table*}

\begin{table*}[ht]\scriptsize
	\caption{ Effectiveness of key modules. All models are trained on the PIDray {\it training} subset and tested on the PIDray {\it test} set. CA, SA and DR represent channel attention, spatial attention and dependency refinement respectively. The accuracies are denoted by ``detection AP/segmentation AP''.}
	\label{tab:detection different parts}
	\begin{center}
		\renewcommand{\arraystretch}{1.2}
		\setlength{\tabcolsep}{1.3mm}
		\begin{tabular}{ccc|cccccccc}
			\hline
			    CA     &     SA     &     DR     &     AP      & $\text{AP}_{50}$ & $\text{AP}_{75}$ & $\text{AP}_{S}$ & $\text{AR}_{1}$ & $\text{AR}_{10}$ & $\text{AR}_{100}$ & $\text{AR}_{S}$ \\ \hline
			           &            &            & $65.4/52.7$ &   $80.3/77.6$    &   $72.8/60.5$    &   $71.4/57.5$   &  $68.0/55.9 $   &   $71.7/58.2$    &    $71.7/58.2$    &   $71.7/58.2$   \\
			\checkmark &            &            & $65.7/52.7$ &   $80.9/77.9$    &   $73.1/60.4$    &   $71.6/57.5$   &   $68.2/55.9$   &   $71.9/58.1$    &    $71.9/58.1$    &   $71.9/58.1$   \\
			           & \checkmark &            & $65.9/53.2$ &   $81.0/78.2$    &   $73.2/61.0$    &   $71.9/58.1$   &   $68.6/56.5$   &   $72.3/58.8$    &    $72.3/58.9$    &   $72.3/58.9$   \\
			\checkmark & \checkmark &            & $66.1/53.1$ &   $80.9/78.3$    &   $73.8/61.0$    &   $72.1/58.0$   &   $68.8/56.4$   &   $72.4/58.6$    &    $72.4/58.6$    &   $72.4/58.6$   \\
			\checkmark & \checkmark & \checkmark & $66.6/53.4$ &   $81.7/78.6$    &   $74.3/61.4$    &   $72.6/58.3$   &   $69.2/56.6$   &   $73.0/58.9$    &    $73.0/58.9$    &   $73.0/58.9$   \\ \hline
		\end{tabular}
	\end{center}
\end{table*}

To verify the design of our proposed Dense Attention Modules (DAMs) in the fine-grained node, we conduct a number of experiments. We use the Cascade Mask R-CNN \cite{cai2019cascade} with ResNet-101-FPN backbone and our divide-and-conquer pipeline as baselines. The experimental performance is presented in Table \ref{tab:detection different parts}. We can see that different modules in our method improve the baseline strikingly when adding them one by one. Desirably, the assembly of these modules contributes to better performance. We believe that it is because our proposed DAMs not only fuse different feature maps in the FPN at each scale depending on their different importance, but also solve the problem of long distance dependency between different pixels to a certain extent.

\begin{table}[h!]
	\caption{Comparison with contribution of different parts of network}
	\label{tab:different parts}
	\begin{center}
		\renewcommand{\arraystretch}{1.2}
		\setlength{\tabcolsep}{3mm}
		\begin{tabular}{c | c }
			\hline
			           Method            & mAP   \\ \hline
			       Baseline (MAX)        & 91.91 \\
			       Baseline (GAP)        & 92.43 \\
			   Cross-Attention Module    & 92.79 \\
			+Divide-and-Conquer Pipeline & 93.16 \\ \hline
		\end{tabular}
	\end{center}
\end{table}

\subsubsection{contributions of fine-grained node to multi-label classification task}

To demonstrate the task-specific contribution in terms of the multi-label classification, we conduct ablation experiments with different components of the network on the top of CvT-21-384 backbone. In the first line of Table \ref{tab:different parts}, we use global average pooling (GAP) to process the feature map $\mathcal{F}$ and then deliver it to classifier directly. In the second line, we adopt $\mathcal{C}$ different queries to obtain  $\mathcal{C}$ class-aware feature vectors, which are used to finalize $\mathcal{C}$ binary classifications. As can be seen from Table \ref{tab:different parts}, it proves the effectiveness of our designed module. Next, we introduce the divide-and-conquer pipeline, which brings an improvement of 0.37 in terms of mAP, demonstrating that divide-and-conquer pipeline is conducive to multi-classification task as well.

\section{Conclusion}
\label{conclution}
In this paper, we construct a challenging dataset (namely PIDray) for prohibited item detection, especially dealing with the cases that the prohibited items are hidden in other objects. PIDray is the largest prohibited item detection dataset so far to our knowledge. Moreover, all images with prohibited items are annotated with bounding boxes and masks of instances. Meanwhile, we design a divide-and-conquer pipeline to make the proposed model more suitable for the real-word application. Specifically, we adopt the tree-like structure to suppress the influence of long-tailed issue in the PIDray dataset, where the first course-grained node is tasked with a binary classification to alleviate the influence of head category, while the subsequent fine-grained node is dedicated to the specific tasks of the tail categories. Based on this simple yet effective scheme, we offer strong task-specific baselines across the object detection and instance segmentation to multi-label classification task on the PIDray dataset and verify its generalization ability on common datasets (like COCO and PASCAL VOC). We hope that the proposed dataset will help the community to establish a unified platform for evaluating the prohibited item detection methods towards real applications. For future work, we plan to extend the current dataset to include more images as well as richer annotations for comprehensive evaluation.

\bibliographystyle{sn-basic}
\bibliography{reference}


\end{document}